\documentclass[runningheads]{llncs}

\usepackage{eccv}

\usepackage{eccvabbrv}

\usepackage{graphicx}
\usepackage{booktabs}

\makeatletter
\def\thanks#1{\protected@xdef\@thanks{\@thanks
		\protect\footnotetext{#1}}}
\makeatother

\usepackage[accsupp]{axessibility}  % Improves PDF readability for those with disabilities.

\usepackage{fontawesome}
\usepackage{multirow}
\usepackage{colortbl}
\usepackage{float}

\usepackage{wrapfig}
\usepackage{marvosym}
\usepackage{pifont}
\newcommand{\cmark}{\ding{51}}
\newcommand{\xmark}{\ding{55}}

\def\etal{\textit{et al.}}

\def\ie{i.e.}
\definecolor{Gray}{gray}{0.9}

\usepackage{hyperref}

\usepackage{orcidlink}

\begin{document}

% ---------------------------------------------------------------
% TODO REVIEW: Replace with your title
\title{LoViF 2026 The First Challenge on Unified Removal of Raindrops and Reflections: Methods and Results} 

% TODO REVIEW: If the paper title is too long for the running head, you can set
% an abbreviated paper title here. If not, comment out.
\titlerunning{LoViF 2026 Challenge on UR$^3$}

% TODO FINAL: Replace with your author list. 
% Include the authors' OCRID for the camera-ready version, if at all possible.
\author{Zewei He\textsuperscript{$\dagger$,\Letter}\thanks{\textsuperscript{$\dagger$} Equal contribution. Z.~He, X.~Tong, Y.~Chen, X.~Liu, and X.~Li are the challenge organizers. The other authors are participants of the LoViF 2026 Challenge on Unified Removal of Raindrops and Reflections.}\thanks{\textsuperscript{\Letter} Corresponding author (zeweihe@zju.edu.cn).}\orcidlink{0000-0003-4280-9708} \and
	Xi Tong\textsuperscript{$\dagger$}\orcidlink{0000-0001-6149-3652} \and
	Yu Chen\textsuperscript{$\dagger$}\orcidlink{0009-0002-6066-6990} \and
	Xingyu Liu\textsuperscript{$\dagger$}\orcidlink{0009-0003-3815-4953} \and
	Xin Li\textsuperscript{$\dagger$}\orcidlink{0000-0002-6352-6523} \and
	Zepeng Wang \and
	Jiagao Hu \and
	Fuhao Li \and
	Yuxuan Chen \and
	Fei Wang \and
	Daiguo Zhou	\and
	Minmin Yi \and
	Chuanrui Zhang \and
	Liwen Zhang \and
	Yeongjin Jeong \and
	Hyunjin Cho \and
	Jiwon Lee \and
	Minsang Kim \and
	Jae Woong Soh \and
	Jin-Hui Jiang \and
	Rong-Lin Jian \and
	Chih-Chung Hsu \and
	Youngjin Oh \and
	Junhyeong Kwon \and
	Junyoung Park \and
	Jae Hyun Park \and
	Sung Ju Lee \and
	Nam Ik Cho \and
	Vishwajeet Shukla \and
	Himanshu Baurai \and
	Zhiqi Zhang \and
	Kui Jiang \and
	Zhaocheng Yu \and
	Runzhe Li \and
	Dawei Fan \and
	Hao Li \and
	Zhanshuo Zhang \and
	Fan Ji \and
	Jiangmeng Li \and
	Xiongxin Tang \and
	Fanjiang Xu \and
	Shangquan Sun \and
	Anh-Kiet Duong \and
	Petra Gomez-Kr\"amer \and
	Jean-Michel Carozza \and
	Ruibo Zhang \and
	Dexiang Hong \and
	Xinyan Liu \and
	Shengeng Tang \and
	Weidong Chen \and
	Tzu-Hsuan Weng \and
	Min-Te Sun
	}

% TODO FINAL: Replace with an abbreviated list of authors.
\authorrunning{Z.~He et al.}
% First names are abbreviated in the running head.
% If there are more than two authors, 'et al.' is used.

% TODO FINAL: Replace with your institution list.
\institute{
	\email{zeweihe@zju.edu.cn}\\
	\textbf{Data page}: \url{https://github.com/hezw2016/RDRF-dataset}
}

\maketitle

\setcounter{footnote}{0} % 正文注脚重新从 1 开始

%%%%%%%%%%%%%%%%%%%%%%%%%%%%%%%%%%%%%%%%%%%%%%%%%%%%%%%%%%%%%%%%%%%%%%%%%%%%%%%%%%%%%%%%%%%%%%%%%%%%%%%%%%%%%%%%%%%%%%%%%%%

\begin{abstract}
This workshop paper comprehensively reviews the First Challenge on Unified Removal of Raindrops and Reflections.
The challenge aims to address a frequently encountered practical problem in the field of autonomous driving, \ie, raindrop-reflection composite degradation on rainy days.
This competition attracted 149 registered participants and received 12 valid final submissions with corresponding fact sheets, significantly contributing to the progress of unified removal of raindrops and reflections.
All the methods are developed and evaluated on our real-shot RainDrop and ReFlection (RDRF) dataset. 
A detailed analysis of the submitted methods and corresponding results is provided in this report, which highlights effective approaches and provides interesting insights for future research.

\keywords{Raindrop removal \and Reflection removal \and Challenge}
\end{abstract}

%%%%%%%%%%%%%%%%%%%%%%%%%%%%%%%%%%%%%%%%%%%%%%%%%%%%%%%%%%%%%%%%%%%%%%%%%%%%%%%%%%%%%%%%%%%%%%%%%%%%%%%%%%%%%%%%%%%%%%%%%%%
\section{Introduction}

On rainy days, raindrops and reflections frequently co-occur in autonomous driving scenarios, posing challenges for onboard visual recognition systems or vehicle cameras during recording.
Adherent raindrops and the reflections from camera side significantly reduce the visibility of captured images~\cite{You2016TPAMI}, and may lead to severe driving safety hazards.
Therefore, \textbf{U}nified \textbf{R}emoval of \textbf{R}aindrops and \textbf{R}eflections (UR$^3$) is a practical problem that requires urgent solution.

Unified removal of raindrops and reflections (UR$^3$) aims to develop a unified model that can handle the raindrop-reflection composite degradation.
Previously, researchers treated raindrop removal and reflection removal as two separate tasks ~\cite{Qian2018CVPR-AGAN,Quan2019ICCV,Shao2021TIP-UMAN,Hu2024NeurIPS-DSIT,Zhao2025CVPR-RDNet,Hu2026AAAI-DAI}.
Though these methods can achieve relatively good performance in removing the target type of degradation (\ie, raindrop or reflection) from a single image, they often fail to remove both types at the same time.

To facilitate the research in this direction, we organized the First Challenge on Unified Removal of Raindrops and Reflections in conjunction with the Second Low-level Vision Frontiers (LoViF) Workshop at ECCV 2026.
This challenge aims to find a practical solution capable of simultaneously eliminating this raindrop-reflection composite degradation, thereby enhancing the clarity of captured images.
We hope this endeavor can provide useful support for applications such as autonomous driving, photography, and video surveillance~\cite{Chen2024ECCV-PTTD,Chen2024TIP-DEANet}.

%This challenge is one of the challenges associated with the LoViF Workshop at ECCV 2026.
%There are a series of challenges on: real-world all-in-one image restoration, unified removal of raindrops and reflections (\textbf{ours}), ultra-low bitrate image compression, day and night raindrop removal for dual-focused images, AIGC image compression, action-conditioned physical and causal world modeling for embodied AI.
This challenge is held as part of the LoViF Workshop, which hosts a series of related challenges, including the Second Real-World All-in-One Image Restoration Challenge~\cite{lovif2026allinone}, the Unified Removal of Raindrops and Reflections Challenge (\textbf{ours}), the Ultra-Low Bitrate Image Compression Challenge~\cite{lovif2026ultralow}, the AIGC Image Compression Challenge~\cite{lovif2026aigc}, the Day and Night Raindrop Removal for Dual-Focused Images Challenge~\cite{lovif2026raindrop}, and the ActPhysCause Challenge~\cite{lovif2026actphyscause}.

\section{UR$^3$ Challenge}
This UR$^3$ challenge is the first competition organized to advance the development of unified removal of raindrops and reflections.
The details of the whole challenge are as follows:

\subsection{RDRF dataset}
The dataset used in this challenge is the RainDrop and ReFlection (RDRF) dataset \cite{Liu2026ECCV-DiffUR3}, which is captured with a dedicated hardware system.
In total, the RDRF dataset consists of 307 unique scenes.
During this challenge, it is divided into a training set (216 scenes with 9003 image pairs) and a validation set (91 scenes with 277 image pairs).
The above images are shared with the research community under a CC BY-NC-SA 4.0 copyright license on \url{https://github.com/hezw2016/RDRF-dataset}.
%
%The RDRF dataset covers a diverse range of urban scene categories with varying semantic attributes.
%As shown in Fig.~\ref{fig:sm-fig1}, the category distribution diagram reveals a rich diversity of semantic groups, along with a relatively balanced proportion across them, thereby providing a solid foundation for the UR$^3$ challenge.
%
%
%\begin{figure}[t]
%	\centering
%	\includegraphics[width=0.5\linewidth]{sm-fig-sunburst}
%	\caption{The category distribution diagram illustrates the hierarchical structure of our dataset. The inner ring represents the distribution of eight primary categories and their proportions, while the outer ring showcases the corresponding representative scenarios for each category.}
%	%	\vspace{-25pt}
%	\label{fig:sm-fig1}
%\end{figure}

In addition, we also captured 100 image pairs from 55 scenes as the testing set for this challenge. To ensure the fairness of this competition, only the degraded images were publicly released.

\subsection{Evaluation protocol}
For all the submitted methods, we followed the scoring function from \cite{Li2025CVPRW} to determine the ranking.
It involves three widely-used metrics, \ie, PSNR, SSIM and LPIPS.
The final score is calculated by re-weighting them as:
\begin{equation}
	\text{Score} = \text{PSNR}(\text{Y})+10\times \text{SSIM}(\text{Y})-5\times \text{LPIPS}.
\end{equation}

For PSNR and SSIM, the evaluation is conducted on the Y channel of the YCbCr color space.
%after converting the restored and the ground-truth images from the RGB color space to YCbCr.
For LPIPS, the image pixel values are normalized to the range of $[-1,1]$ before computing the perceptual distance (based on AlexNet) between the restored and the ground-truth images.
\begin{table}[t]
	\footnotesize
	\centering
	\caption{Quantitative results on the UR$^3$ Challenge test set. The \textbf{bold} and \underline{underline} indicate the best and second-best results.}
	\resizebox{0.9\linewidth}{!}{
		\begin{tabular}{ccc|ccc|c|c}
			\toprule[1.2pt]
			Rank & Team & Leader & PSNR$\uparrow$ & SSIM$\uparrow$ & LPIPS$\downarrow$ & Extra Data & Score$\uparrow$ \\
			\midrule \midrule
			1  & \underline{X}edit \underline{M}aster     & Jiagao Hu          & \textbf{32.3993} & 0.9645 & 0.0443 & \cmark & \textbf{41.8226} \\
			2  & tysl              & Minmin Yi          & \underline{32.3006} & \underline{0.9648} & \underline{0.0416} & \cmark & \underline{41.7405} \\
			3  & GIST-IVL          & Yeongjin Jeong     & 31.6556 & \textbf{0.9680} & \textbf{0.0388} & \cmark & 41.1416 \\
			4  & ACVLAB            & Jin-Hui Jiang      & 31.0060 & 0.9603 & 0.0432 & \xmark & 40.3933 \\
			5  & SNU-ISPL          & Youngjin Oh        & 30.7389 & 0.9572 & 0.0526 & \cmark & 40.0474 \\
			6  & the\_last\_token  & Vishwajeet Shukla  & 30.3074 & 0.9580 & 0.0559 & \cmark & 39.6075 \\
			7  & AIIA-Lab          & Zhiqi Zhang        & 30.3203 & 0.9568 & 0.0563 & \xmark & 39.6067 \\
			8  & ISCAS Optics     & Dawei Fan          & 30.1058 & 0.9575 & 0.0611 & \xmark & 39.3752 \\
			9  & sunsean           & Shangquan Sun      & 29.7094 & 0.9562 & 0.0652 & \cmark & 38.9455 \\
			10 & ULR               & Anh-Kiet Duong     & 29.0555 & 0.9260 & 0.1107 & \xmark & 37.7620 \\
			11 & ustc\_pi\_lab     & Ruibo Zhang        & 28.2417 & 0.9457 & 0.0851 & \xmark & 37.2730 \\
			12 & daniel0902        & Tzu-Hsuan Weng     & 27.8509 & 0.9440 & 0.0800 & \xmark & 36.8909 \\
			\bottomrule[1.2pt]
	\end{tabular}}
	\label{tab:challenge_results}
\end{table}

\begin{table}[t]
	\footnotesize
	\centering
	\caption{Quantitative results on the UR$^3$ Challenge test set. The results are obtained with submitted inference codes under strict single-frame setting. The \textbf{bold} and \underline{underline} indicate the best and second-best results.}
	\resizebox{0.9\linewidth}{!}{
		\begin{tabular}{ccc|ccc|c|c}
			\toprule[1.2pt]
			Rank & Team & Leader & PSNR$\uparrow$ & SSIM$\uparrow$ & LPIPS$\downarrow$ & Extra Data & Score$\uparrow$ \\
			\midrule \midrule
			1 & \underline{X}edit \underline{M}aster & Jiagao Hu & \textbf{32.3993} & \underline{0.9645} & 0.0443 & \cmark & \textbf{41.8226} \\
			2 & tysl & Minmin Yi & \underline{32.3006} & \textbf{0.9648} & \textbf{0.0416} & \cmark & \underline{41.7405} \\
			3 & ACVLAB & Jin-Hui Jiang & 31.0060 & 0.9603 & \underline{0.0432} & \xmark & 40.3933 \\
			4 & SNU-ISPL & Youngjin Oh & 30.7389 & 0.9572 & 0.0526 & \cmark & 40.0474 \\
			5 & GIST-IVL$^\dagger$ & Yeongjin Jeong & 30.5155 & 0.9573 & 0.0516 & \cmark & 39.8308 \\
			6 & the\_last\_token & Vishwajeet Shukla & 30.3074 & 0.9580 & 0.0559 & \cmark & 39.6075 \\
			7 & ISCAS Optics & Dawei Fan & 30.1058 & 0.9575 & 0.0611 & \xmark & 39.3752 \\
			8 & sunsean & Shangquan Sun & 29.7094 & 0.9562 & 0.0652 & \cmark & 38.9455 \\
			9 & AIIA-Lab$^\dagger$ & Zhiqi Zhang & 29.7247 & 0.9544 & 0.0657 & \xmark & 38.9404 \\
			10 & ustc\_pi\_lab & Ruibo Zhang & 28.2417 & 0.9457 & 0.0851 & \xmark & 37.2730 \\
			11 & daniel0902 & Tzu-Hsuan Weng & 27.8509 & 0.9440 & 0.0800 & \xmark & 36.8909 \\
			12 & ULR$^\dagger$ & Anh-Kiet Duong & 28.2267 & 0.9215 & 0.1176 & \xmark & 36.8534 \\
			\bottomrule[1.2pt]
		\end{tabular}
	}
	\label{tab:challenge_results2}
\end{table}

%and then the perceptual distance is computed using the AlexNet-based configuration between the restored and the ground-truth images

\subsection{Challenge phases}
There are two phases in this challenge, \ie, the development and testing phases. The details are as follows:

\subsubsection{Development phase:} In this phase, we released the training set in our RDRF dataset. The degraded images in the validation set were also released. Each participant can upload their restored images to the challenge platform to obtain the corresponding score on the validation set. During this phase, we received 1176 submissions from 56 participants. 

\subsubsection{Testing phase:} In this phase, we released the ground-truth images in the validation set. The degraded images in the testing set were also released. Each participant can upload their restored images to the challenge platform to obtain the corresponding score on the testing set. During this phase, we received 406 submissions from 38 participants.

\section{Challenge Results}
We have summarized the challenge results in Table~\ref{tab:challenge_results}.
Team \textbf{\underline{X}edit \underline{M}aster} achieved the best overall performance in the challenge, ranking top with a final score of 41.8226.
This team also achieved the highest PSNR value (32.3993).
Team \textbf{tysl} ranked second with a final score of 41.7405.
This team secured the second position across three indicators (\ie, PSNR, SSIM and LPIPS).
Team \textbf{GIST-IVL} ranked third with a final score of 41.1416.
This team also achieved the highest SSIM and LPIPS values (0.9680 and 0.0388).

Due to the acquisition protocol of the RDRF dataset, a single ground-truth image may correspond to multiple degraded images.
Consequently, multi-frame fusion can enhance restoration performance.
We found that some submitted methods utilized multiple images of the same scene to produce the final output, which was somewhat unfair compared with single-frame methods.
This challenge was originally designed as a single-image restoration task. 
Therefore, the inference should rely on only one input image. 
Using multiple images or frames is inconsistent with the task setting and does not match the intended practical scenario.

We re-tested the results with submitted inference codes under strict single-frame setting.
Table~\ref{tab:challenge_results2} shows the results.
Team \textbf{ACVLAB} ranked third by removing multi-frame fusion.

\section{Teams and Methods}

\subsection{\underline{X}edit \underline{M}aster}

\begin{figure}[t]
	\centering
	\includegraphics[width=0.65\linewidth]{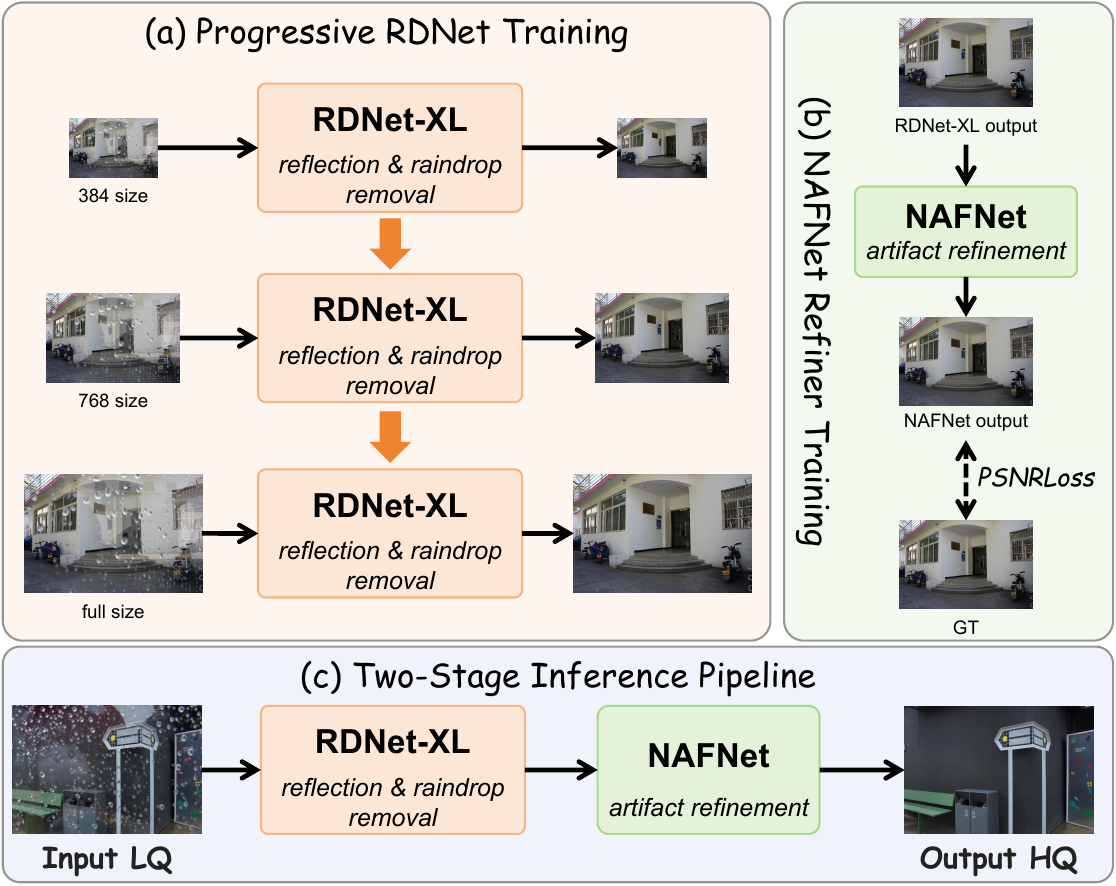}
	\caption{The pipeline of the method proposed by Team \underline{X}edit \underline{M}aster.}
	\label{fig:xedit-master}
\end{figure}

This team proposes Progressive Cascaded Removal and Refinement, a two-stage framework built upon an RDNet remover and a NAFNet refiner, as illustrated in Fig.~\ref{fig:xedit-master}. In the first stage, the remover jointly suppresses raindrops and reflections. In the second stage, the refiner corrects the remaining artifacts and recovers fine image details. By separating degradation removal from fidelity-oriented refinement, the method allows the two networks to be optimized for complementary objectives.

The first-stage remover adopts RDNet~\cite{Zhao2025CVPR-RDNet} with a larger FocalNet-XL backbone~\cite{Yang2022NeurIPS-FocalNet}. Building upon the trained remover, the second stage adopts the SIDD configuration of NAFNet~\cite{Chen2022ECCV-NAFNet}, with a width of 64 and initialization from SIDD-pretrained weights. Specifically, the team freezes RDNet-XL and applies it to the entire training set to obtain intermediate restorations. These results are used as the inputs to NAFNet, while the corresponding clean images serve as supervision. The refiner is trained for 200 epochs using a PSNR-based loss and a cosine-annealing learning-rate schedule to correct residual artifacts remaining after the first-stage restoration.

\noindent \textbf{Training Details.} The models are trained sequentially. The team first follows the three-resolution curriculum to optimize RDNet-XL and then trains NAFNet using the frozen RDNet-XL outputs. In addition to the official challenge training data, they use the DeRaindrop~\cite{Qian2018CVPR-AGAN} and RainDS~\cite{Quan2021CVPR-CCN} datasets for raindrop removal, together with OpenRR-5K~\cite{Cai2025MIPR-OpenRR} and RR4K for reflection removal.

\noindent \textbf{Testing Details.} During testing, each degraded image is first processed by RDNet-XL for unified raindrop and reflection removal. The resulting image is then passed once through NAFNet to obtain the final restoration.

\noindent \textbf{Implementation Details.} The method is implemented on eight NVIDIA A100 GPUs using bf16 mixed precision. The remover adopts the FocalNet-XL version of RDNet, while the refiner adopts the NAFNet architecture configured with a base channel width of 64 following its standard SIDD denoising setting.

\begin{figure*}[t]
	\centering
	\includegraphics[width=0.75\textwidth]{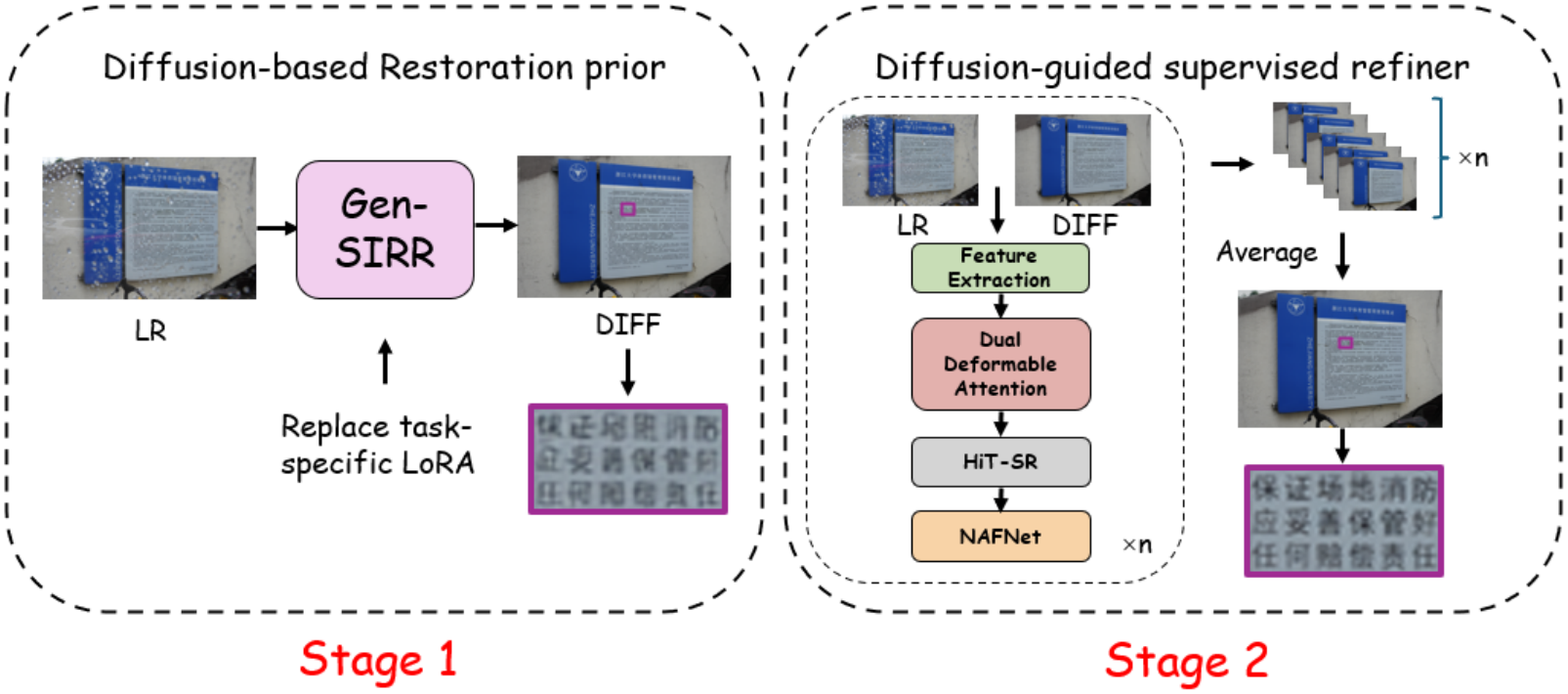}
	\caption{The pipeline of the method proposed by Team tysl.}
	\label{fig:tysl}
\end{figure*}

\subsection{tysl}

This team proposes a diffusion-guided two-stage framework that combines a generative clean-image prior with a supervised fidelity-oriented refiner, as illustrated in Fig.~\ref{fig:tysl}. In the first stage, they adapt the pretrained GenSIRR model~\cite{Li2026CVPR-GenSIRR} to the UR$^3$ task using LoRA~\cite{Hu2022ICLR-LoRA}. Specifically, a task-oriented prompt guides the diffusion model to remove reflections and raindrops while completing the content occluded by raindrops. Since the generated result may contain local geometric or color inconsistencies, the method uses it as a restoration prior instead of directly treating it as the final output.

In the second stage, a supervised refiner takes the original degraded image and the diffusion prior as inputs. Two shallow convolutional branches first extract their features, which are aligned and exchanged through bidirectional deformable attention~\cite{Zhu2020ArXiv-DeformDETR} to correct local inconsistencies in the diffusion prior. A ConvNeXt-based estimator~\cite{Liu2022CVPR-ConvNeXt} further predicts a low-resolution spatial prompt map to adaptively modulate the refinement features. The prompted features are concatenated and fused by HiT-SR~\cite{Zhang2024ECCV-HiTSR}, after which NAFNet predicts a residual correction that is added to the diffusion prior.

\begin{figure}[t]
	\centering
	\includegraphics[width=0.7\linewidth]{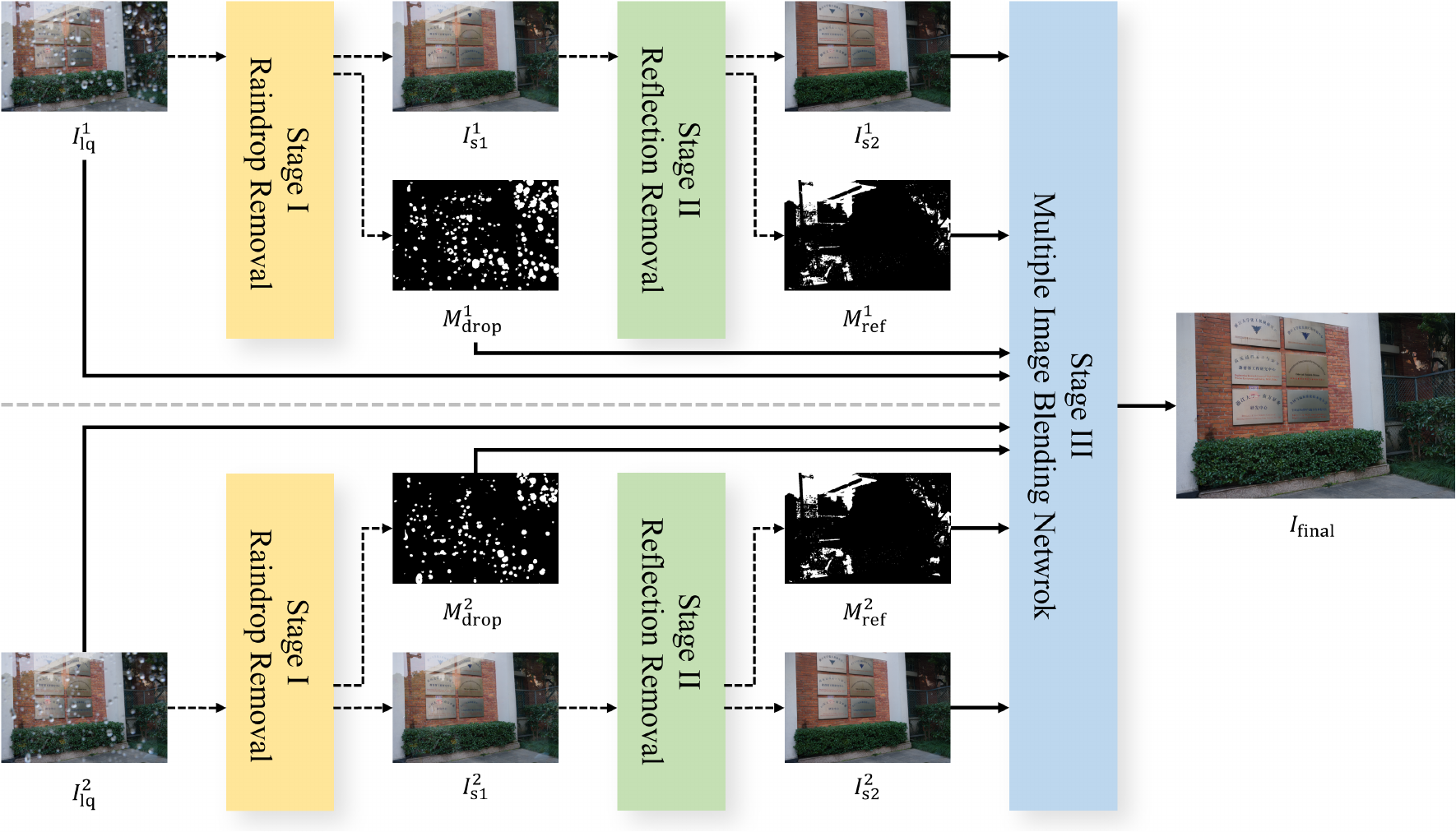}
	\caption{The overall framework of Team GIST-IVL.}
	\label{fig:gist-ivl}
\end{figure}

\noindent \textbf{Training Details.} Training comprises two stages. First, the pretrained GenSIRR model is adapted with LoRA for 20,000 iterations using a task-specific restoration prompt, a learning rate of $1\times10^{-4}$. The refiner is then trained for 60,768 iterations, taking the degraded image and the GenSIRR-generated restoration prior as inputs and the corresponding clean image as the target.

\noindent \textbf{Testing Details.} During testing, the adapted GenSIRR model uses 24 denoising steps to generate a restoration prior for each test image. The generated prior and the original degraded image are then jointly fed into the trained refiner to produce the restored output.

\noindent \textbf{Implementation Details.} The second-stage refiner is trained on four NVIDIA A100 GPUs, each with 80 GB of memory. The two stages are trained independently. During the first-stage adaptation, only the LoRA parameters are updated, while the pretrained GenSIRR weights remain frozen.

\noindent \textbf{Ensembles and Fusion Strategies.} For each test image, the adapted GenSIRR model is sampled using 20 fixed random seeds, ranging from 1 to 20. Each generated restoration prior is independently processed by the refiner, and the resulting 20 restored outputs are averaged to obtain the final prediction.

\subsection{GIST-IVL}

\begin{figure}[t]
	\centering
	\includegraphics[width=0.8\linewidth]{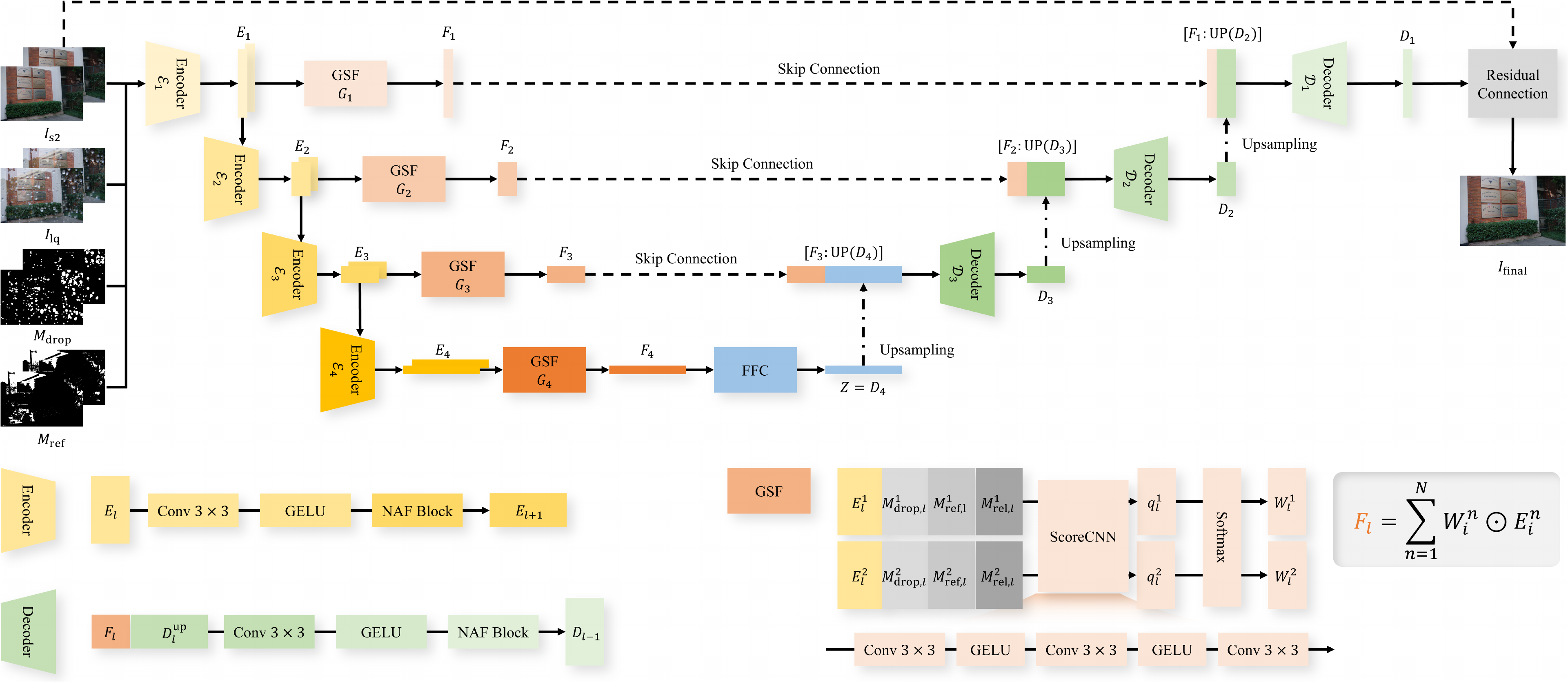}
	\caption{The architecture of Stage III in MAFNet-UR$^3$.}
	\label{fig:gist-ivl-stage3}
\end{figure}

Team GIST-IVL proposes MAFNet-UR$^3$, a three-stage framework that exploits multiple observations of the same scene for unified raindrop and reflection removal, as illustrated in Fig.~\ref{fig:gist-ivl}. In Stage I, a pretrained WeatherRemover model~\cite{Qu2025TAI-WeatherRemover} is fine-tuned with raindrop-free targets that retain reflections, producing an intermediate result for subsequent reflection removal. Building on this output, Stage II adapts FUMO~\cite{Xu2026ArXiv-FUMO} to the UR$^3$ task. An intensity prior is generated by Qwen3-VL~\cite{Bai2025ArXiv-Qwen3VL}, while a high-frequency prior is extracted through wavelet decomposition. Conditioned on these complementary cues, a Stable Diffusion 2.1-based model~\cite{Rombach2022CVPR-LDM} with a ControlNet-style branch~\cite{Zhang2023ICCV-ControlNet} performs one-step restoration. The preliminary result is then refined by NAFNet using the two priors and the Stage I output as guidance. Stage III, illustrated in Fig.~\ref{fig:gist-ivl-stage3}, integrates the input with the preceding stage outputs. Raindrop and reflection masks are estimated from their luminance differences, after which a shared NAFBlock encoder extracts multi-scale features and gated spatial fusion modules assign mask-aware weights. The fused features are further processed by Fast Fourier Convolution blocks~\cite{Chi2020NeurIPS-FFC} and reconstructed by an anchor-based residual decoder, suppressing corrupted regions while preserving reliable structures from cleaner observations.

\noindent \textbf{Training Details.} Training follows the three-stage pipeline. Stage I is optimized for 50k iterations using Adam and a Charbonnier loss. In Stage II, the diffusion model and refiner are trained for 200k and 25k iterations, respectively, using the official UR$^3$ training data and Stage I outputs. Stage III is trained for 100k iterations on $256\times256$ patches.

\noindent \textbf{Testing Details.} At inference, Stage I first suppresses raindrops while preserving reflection cues for subsequent processing. Stage II then generates the required priors and performs restoration at $1080\times720$ resolution using bf16 precision. Finally, Stage III jointly processes the grouped observations and produces the final result by predicting a residual correction over a weighted anchor.

\noindent \textbf{Implementation Details.} Stages I, II, and III contain 24.32M, 1.873B, and 38.14M parameters, respectively. Measured on an NVIDIA GeForce RTX 5090, the corresponding runtimes are 1.007, 4.843, and 0.764 seconds per image.

\noindent \textbf{Ensembles and Fusion Strategies.} Stage II employs an 8-way D4 self-ensemble, and Stage III further combines an internal 8-way D4 ensemble with learned mask-aware fusion across multiple observations of the same scene.

\noindent \textbf{Note.} MAFNet-UR$^3$ uses multiple observations of the same scene during inference, which differs from the standard single-image setting adopted by the challenge. For transparency and fair comparison, its results under the multi-observation and single-image settings are reported separately in Tables~\ref{tab:challenge_results} and~\ref{tab:challenge_results2}.

\begin{figure*}[t]
	\centering
	\includegraphics[width=0.8\textwidth]{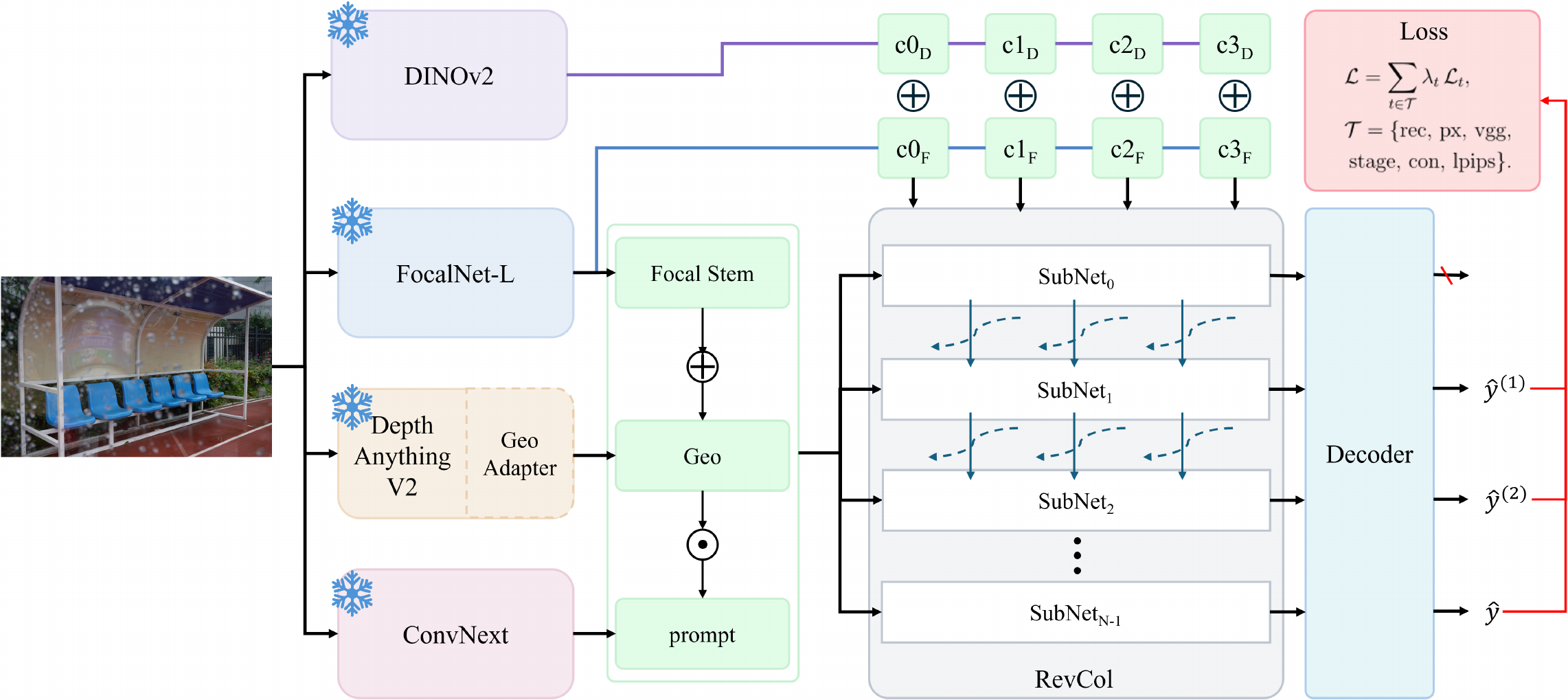}
	\caption{The overall framework of Team ACVLAB.}
	\label{fig:acvlab}
\end{figure*}

\subsection{ACVLAB}

This team proposes RDNet-SGCR, a single feed-forward restoration network illustrated in Fig.~\ref{fig:acvlab}. The method builds upon RDNet by introducing Semantic-and-Geometry-Guided Cascaded Refinement (SGCR). It is motivated by the complementary characteristics of the two degradations: raindrops typically appear as sparse, localized, high-frequency occlusions, whereas reflections are generally low-frequency, semi-transparent layer-mixing artifacts. The backbone uses frozen FocalNet-L and DINOv2~\cite{Oquab2024TMLR-DINOv2} encoders to extract appearance and semantic features, which are fused and modulated by a content-adaptive prompt. A frozen Depth-Anything-V2 model~\cite{Yang2024NeurIPS-DepthAnythingV2} provides depth and surface-normal cues, injected through a zero-initialized geometry adapter to preserve the RDNet initialization. Four reversible sub-networks progressively estimate the restoration residual, with the last three outputs supervised as refinement stages.

\noindent \textbf{Training Details.} Training proceeds in three steps. First, RDNet is initialized from its official reflection-removal checkpoint, equipped with a frozen DINOv2 semantic branch, and fine-tuned on OpenRR-5K~\cite{Cai2025MIPR-OpenRR} for 20 epochs. Second, the resulting backbone is adapted to the RDRF training set using random $640\times640$ crops, Adam with a learning rate of $1\times10^{-4}$ and zero weight decay, a batch size of 1, and mixed precision. Third, the full SGCR model is fine-tuned on $512\times512$ crops with online depth and surface-normal guidance. Its three refinement stages are jointly trained with stage-wise supervision, contraction loss, and final-stage LPIPS loss. Ten percent of the competition data is reserved for validation. EMA, gradient clipping, rain-safe augmentation, and warm-up evaluation are also adopted.

\noindent \textbf{Testing Details.} During testing, depth and surface normals are first estimated from the input and injected through the learned geometry adapter. The resulting features are then processed by the reversible restoration cascade. For high-resolution images, the team optionally adopts overlapping $512\times512$ tiled inference with 25\% overlap and triangular-window blending.

\noindent \textbf{Implementation Details.} The complete pipeline contains approximately 428M parameters, including 267.2M trainable parameters and 160.8M parameters in the frozen encoders. On an NVIDIA A100 40GB GPU, a single forward pass takes 206ms and uses 2.28GiB of peak memory.

\noindent \textbf{Ensembles and Fusion Strategies.} At inference, a four-way rain-safe test-time augmentation (TTA) comprising the identity, $180^{\circ}$ rotation, horizontal flip, and vertical flip is applied, while $90^{\circ}$ and $270^{\circ}$ rotations are excluded to avoid unrealistic horizontal rain streaks. For each transformed input, the depth map and surface normals are recomputed to maintain geometric consistency.

\begin{figure}[t]
	\centering
	\includegraphics[width=0.6\linewidth]{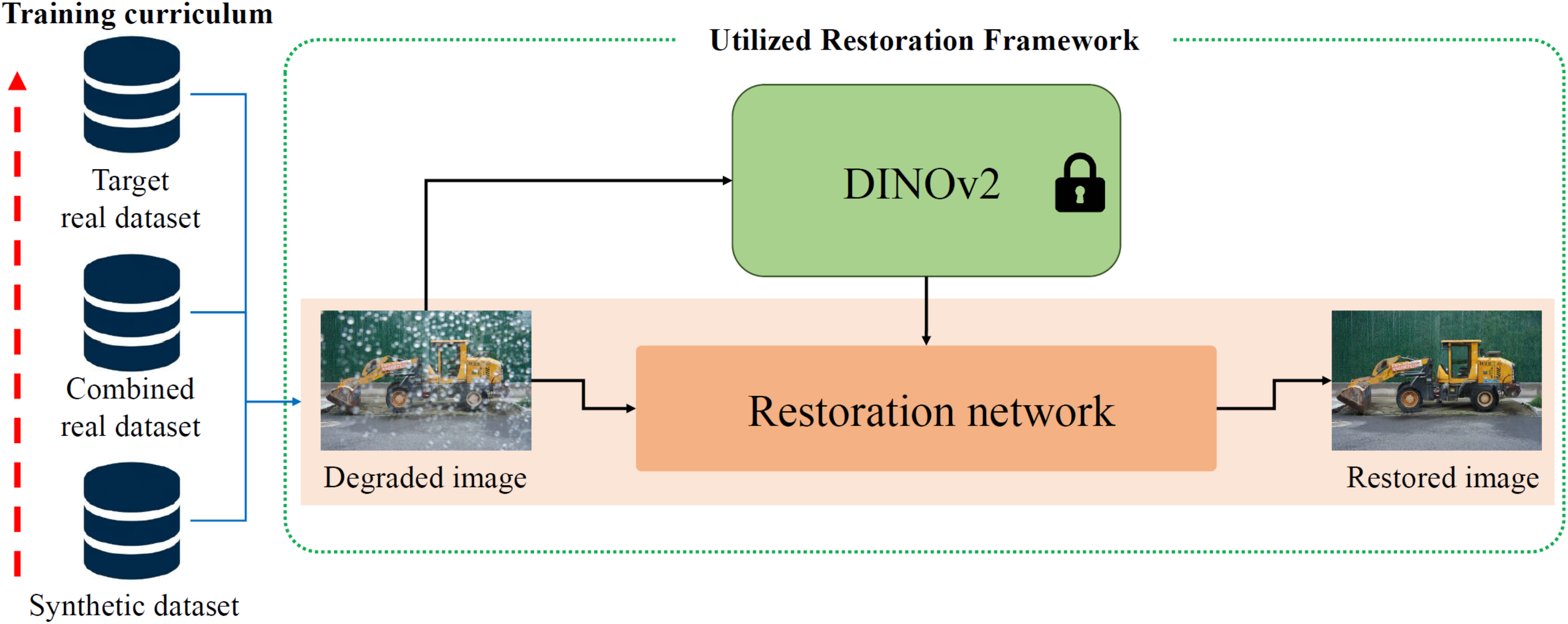}
	\caption{The overall framework of Team SNU-ISPL.}
	\label{fig:snu-ispl}
\end{figure}

\subsection{SNU-ISPL}

This team adopts the hierarchical Transformer architecture of DINOLight~\cite{Oh2026ArXiv-DINOLight} and introduces DINOClear, a data-centric adaptation for unified raindrop and reflection removal, as illustrated in Fig.~\ref{fig:snu-ispl}. Rather than increasing architectural complexity or model capacity, the method improves generalization through a sequential three-phase training curriculum. It further leverages hierarchical self-supervised features from DINOv2 to provide complementary semantic and fine-grained structural priors for distinguishing genuine scene content from raindrop and reflection degradation artifacts. The method employs an Adaptive Feature Fusion Module to combine shallow geometric and structural features with deeper semantic representations from a distilled DINOv2 ViT-B/14 encoder. The fused features are injected into the restoration network through SFDINO blocks, whose Auxiliary Cross-Attention sequentially operates in the spatial and frequency domains to supplement restoration features with self-supervised visual priors.

\noindent \textbf{Training Details.} Training follows a three-phase curriculum: (1) 50k iterations on synthetic reflection pairs generated online from 13,700 source pairs, (2) 50k iterations on 8,138 real reflection pairs collected from OpenRR-5K~\cite{Cai2025MIPR-OpenRR}, Zhang \etal~\cite{Zhang2018CVPR-ReflectionSeparation}, and SIR$^2+$~\cite{Wan2017ICCV}, and (3) 150k iterations on RDRF for target-domain adaptation.

\noindent \textbf{Testing Details.} At inference, high-resolution images are processed using sliding-window inference with $448\times448$ patches and an adjustable overlap ratio.

\noindent \textbf{Implementation Details.} The restoration network contains 9.51M trainable parameters. For a $448\times448$ input, it requires 193.2 GFLOPs excluding the frozen DINOv2 encoder and 369.08 GFLOPs including it.

\noindent \textbf{Ensembles and Fusion Strategies.} The final submission employs an 8-way D4 ensemble, where predictions from transformed inputs are inverse-transformed and averaged. Stochastic weight averaging is also applied after training.

\subsection{the\_last\_token}

\begin{figure*}[t]
	\centering
	\includegraphics[width=0.8\textwidth]{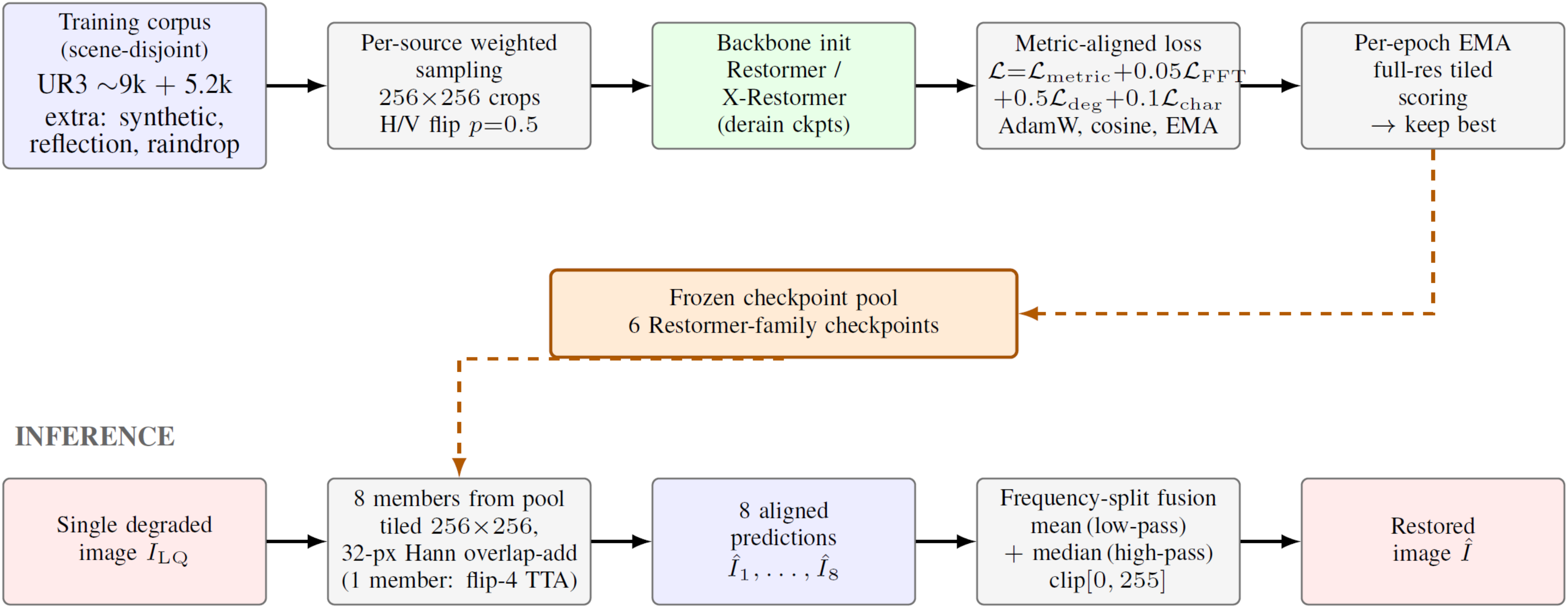}
	\caption{The pipeline of the method proposed by Team the\_last\_token.}
	\label{fig:last-token}
\end{figure*}

This team introduces an ensemble-based restoration framework that combines eight Restormer-family predictions through frequency-split fusion, as illustrated in Fig.~\ref{fig:last-token}. They compare several restoration architectures, including Restormer~\cite{Zamir2022CVPR-Restormer}, X-Restormer~\cite{Chen2023ArXiv-XRestormer}, DRSformer~\cite{Chen2023CVPR-DRS}, InstructIR~\cite{Conde2024ECCV-InstructIR}, and MambaIR~\cite{Guo2024ECCV-MambaIR}, together with the diffusion-prior DAI model~\cite{Hu2026AAAI-DAI}. Based on scene-disjoint validation, Restormer and X-Restormer are selected as the primary backbones. The models are optimized using a differentiable surrogate of the official score, combining negative luma-domain PSNR and SSIM with RGB LPIPS. The objective further includes an FFT-domain reconstruction loss, a degradation-aware weighted $L_1$ loss, and a Charbonnier loss.

\noindent \textbf{Training Details.} Restormer and X-Restormer are initialized from released deraining checkpoints and fine-tuned using AdamW with linear warm-up, cosine decay, gradient clipping, bf16 autocast, and EMA-based model selection. The training set contains 9,003 challenge pairs and 5,161 additional pairs, including 2,975 synthetic joint-degradation pairs generated from Cityscapes~\cite{Cordts2016CVPR-Cityscapes}, 1,018 reflection pairs, and 1,168 raindrop pairs. Online synthetic degradation is further applied to 20\% of the crops.

\noindent \textbf{Testing Details.} Each model performs overlap-add tiled inference using $256\times256$ windows with a 32-pixel overlap, reflection padding, and Hann-window blending. The reconstructed predictions are spatially aligned before fusion.

\noindent \textbf{Implementation Details.} On a single NVIDIA H200 GPU, the per-image runtimes of X-Restormer and Restormer are 1.16 and 0.67 seconds without TTA, and 3.60 and 2.35 seconds with four-way TTA, respectively.

\noindent \textbf{Ensembles and Fusion Strategies.} The final ensemble contains eight predictions generated from a Restormer-family checkpoint pool, with one checkpoint evaluated both with and without four-view orientation-preserving TTA. Low-frequency components are averaged, while high-frequency residuals are fused using a pixel-wise median to suppress inconsistent details.

\subsection{AIIA-Lab}

\begin{figure}[t]
	\centering
	\includegraphics[width=0.75\linewidth]{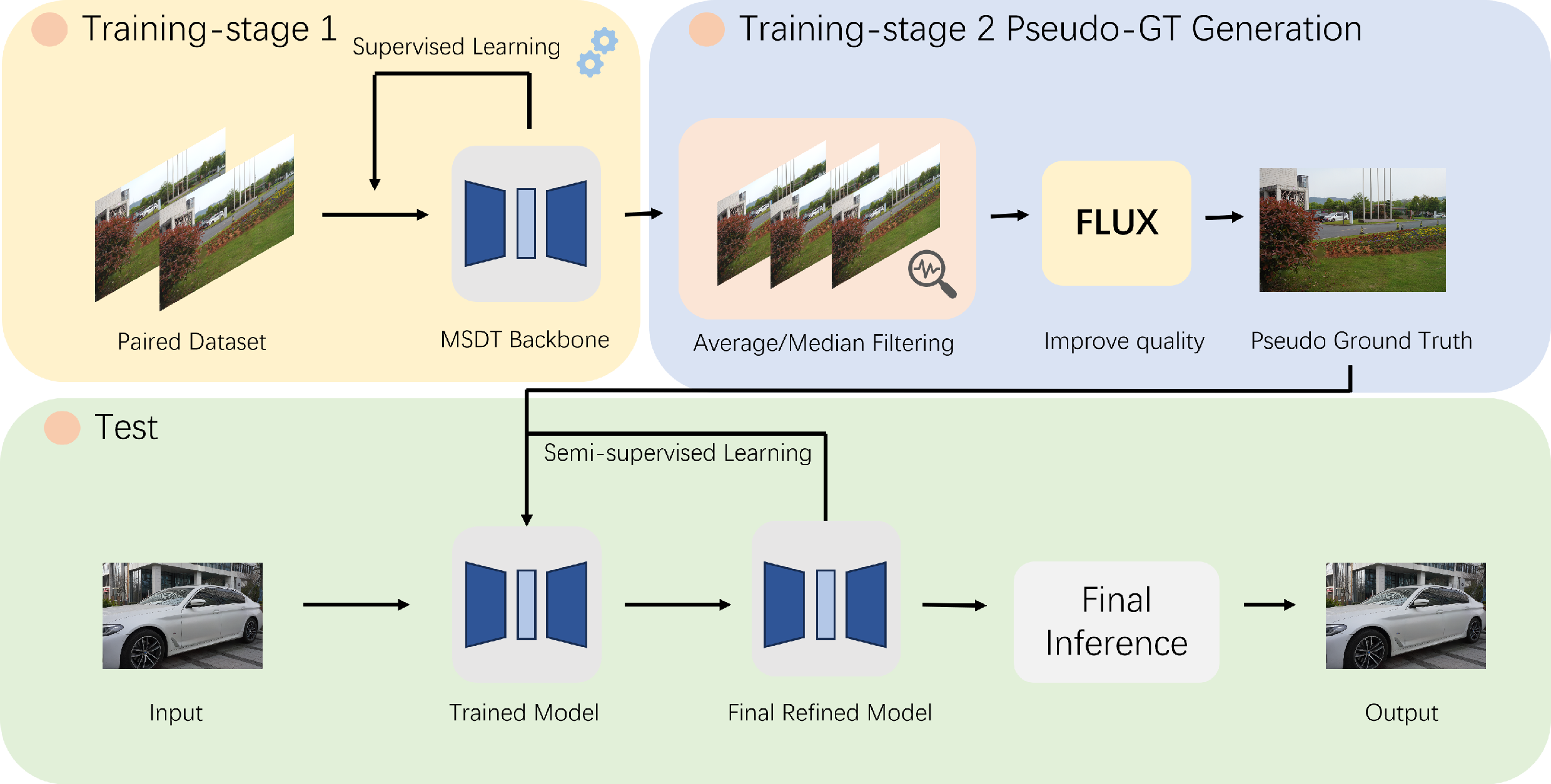}
	\caption{The pipeline of the method proposed by Team AIIA-Lab.}
	\label{fig:aiia}
\end{figure}

The team proposes a multi-stage pipeline for joint raindrop and reflection removal, adopting the Multi-Scale Deraining Transformer (MSDT)~\cite{chen2024AAAI-msdt} 
as the backbone and augmenting it with checkpoint-based scene-specific selection, inter-frames fusion, and Flux~\cite{BlackForestLabs2024-flux} refinement. Its overall pipeline is illustrated in Fig.~\ref{fig:aiia}.

\noindent \textbf{Training Details.}In Stage I, MSDT is trained for 500 epochs on $256\times256$ patches with batch size 1, using Adam and cosine learning-rate decay from $1\times10^{-4}$ to $1\times10^{-6}$ after 3 warm-up epochs. Augmentation includes random cropping, flipping, 90-degree rotations, and gamma and saturation adjustments, while several high-performing checkpoints are retained. In Stage II, their outputs are aggregated by mean or median fusion to form pseudo ground truth, which is then used to fine-tune Flux~\cite{BlackForestLabs2024-flux} as a refinement model.

\noindent \textbf{Testing Details.} At test time, they employ a sliding-window scheme to handle arbitrary resolutions: overlapping predictions are averaged, and edge padding with recomposition addresses non-divisible dimensions. In addition, A set of strong checkpoints of MSDT proceed each test image independently. For each scene, the best-performing checkpoint is selected case-by-case.

\noindent \textbf{Implementation Details.} The team uses MSDT as the primary restoration backbone and a fine-tuned FLUX model for pseudo-GT-based refinement. Both stages are developed exclusively using the official challenge data.

\noindent \textbf{Ensembles and Fusion Strategies.} Outputs of the same scene are fused via mean or median fusion. Mean fusion aggregates shared structural information; median fusion is more robust to local outliers and residual artifacts.

\noindent \textbf{Note.} This team trains its residual correction model using multiple observations of the same scene. Results for the multi-observation and standard single-image settings are reported separately in Tables~\ref{tab:challenge_results} and~\ref{tab:challenge_results2}.

\subsection{ISCAS Optics}
This team uses XResformer \cite{chen2024ECCV-XResformer} as the backbone and proposes a coarse-to-fine training strategy, 
along with a self-ensemble testing strategy.

\noindent \textbf{Training Details.} Training follows a four-stage coarse-to-fine schedule with random cropping, flipping, and rotation augmentation. The model is first trained for 300k iterations on $256\times256$ patches using $\ell_1$ loss, followed by another 300k iterations with $\mathcal{L}_{\ell_1} + 0.01 \times \mathcal{L}_{\text{FFT}}$. The patch size is then increased to $320\times320$ for 210k iterations with $\ell_1$ loss, and finally to $384\times384$ for another 210k iterations using the combined loss. The FFT loss provides frequency-domain supervision for recovering high-frequency details.

\noindent \textbf{Ensembles and Fusion Strategies.} The method combines eight-way geometric self-ensemble with multi-resolution inference. Predictions from rotated and flipped inputs are inversely transformed and averaged. Full-image inference is further fused with sliding-window inference using $384\times384$ patches and a 320-pixel overlap, balancing global context and local detail restoration.

\subsection{sunsean}

This team proposes a two-branch ensemble framework that combines a direct restoration model (RD) with the diffusion-based DAI model~\cite{Hu2026AAAI-DAI}. Given a degraded input image $I_{\mathrm{input}}$, the two branches independently produce restored predictions $I_{\mathrm{RD}}$ and $I_{\mathrm{DAI}}$. The final output is obtained by linear blending. The two branches provide complementary restoration characteristics, with the RD branch performing direct deterministic restoration and the DAI branch providing diffusion-based refinement.

\noindent \textbf{Testing Details.} The RD branch directly processes the degraded image to produce $I_{\mathrm{RD}}$. For high-resolution inputs, the DAI branch adopts overlapping tiled inference using $512\times512$ patches with a 64-pixel overlap. The tile predictions are subsequently merged to reconstruct the full-resolution output.

\noindent \textbf{Ensembles and Fusion Strategies.} The DAI branch uses a scene-adaptive checkpoint selection strategy. For relatively simple scenes, inference is performed using a single checkpoint. For more challenging scenes, two complementary checkpoints independently process the same input, and their predictions are averaged with equal weights. The resulting DAI prediction is then linearly fused with the RD output using the weight defined above.

\subsection{ULR}

This team proposes a conditional diffusion-based restoration framework based on Stable Diffusion 2.1. DINOv3~\cite{simeoni2025-dinov3} features are clustered using Affinity Propagation~\cite{frey2007Science-clustering} to group images with similar degradations and provide cluster-level conditioning. A scene-level pseudo-reference is also generated by pixel-wise median aggregation and concatenated with each input image.

\noindent \textbf{Training Details.} The model is trained for up to 2,000 epochs using Adam with a learning rate of $1\times10^{-5}$ and a batch size of 1. Images are resized to $720\times1080$ and randomly flipped horizontally. The DINOv3 encoder and VAE are frozen, while only the U-Net is updated.

\noindent \textbf{Testing Details.} Each input is concatenated with the scene-level pseudo-reference. Inference is performed in two modes: one uses cluster-level DINOv3 features, while the other replaces them with zeros to reduce dependence on clustering.

\noindent \textbf{Ensembles and Fusion Strategies.} The ensemble consists of two models initialized from the same pretrained weights and trained with different random seeds and data orderings.

\noindent \textbf{Note.} This method uses multiple degraded images from the same scene for conditioning and is therefore evaluated separately from standard single-image methods in Tables~\ref{tab:challenge_results} and~\ref{tab:challenge_results2}.

%\subsection{ustc\_pi\_lab}
%
%This team proposes a two-stage restoration framework combining Histoformer with a signed multi-band residual subtractor, as shown in Fig.~\ref{fig:ustc-pi-lab}. In Stage I, Histoformer~\cite{Sun2024ECCV-Hist} is initialized from a public checkpoint and fine-tuned on the official training set to exploit long-range regions with similar intensity statistics. In Stage II, a residual subtractor estimates the remaining degradation from a 21-channel representation comprising the degraded input, Stage-I output, their difference, low- and high-frequency features, and an in-mask prior.

\subsection{ustc\_pi\_lab}

\begin{figure}[h]
	\centering
	\includegraphics[width=0.7\textwidth]{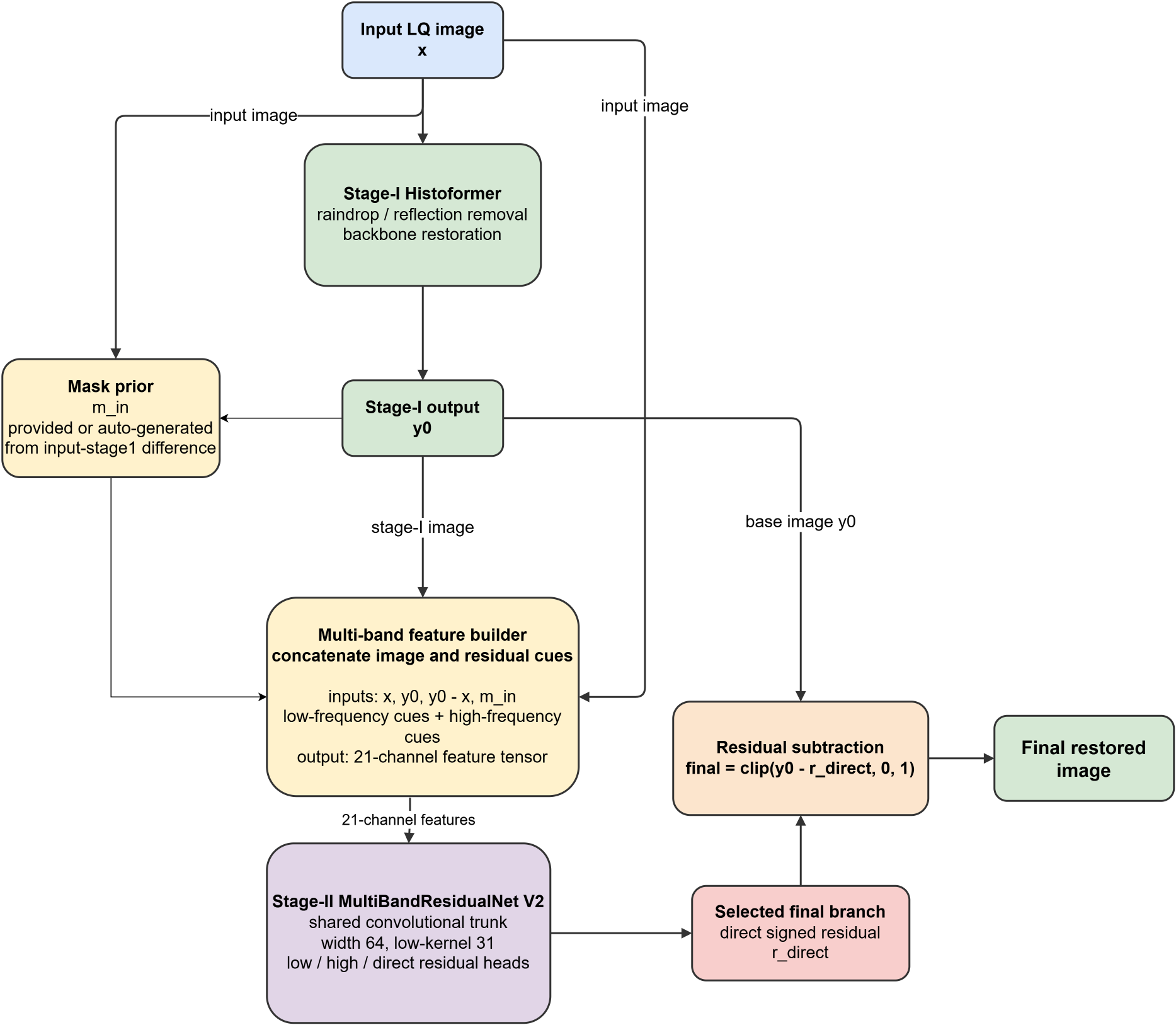}
	\caption{The pipeline of the method proposed by Team ustc\_pi\_lab.
	}
	\label{fig:ustc-pi-lab}
\end{figure}

This team proposes a two-stage restoration framework that combines Histoformer with a signed multi-band residual subtractor, as shown in Fig.~\ref{fig:ustc-pi-lab}. In Stage I, Histoformer~\cite{Sun2024ECCV-Hist} serves as the restoration backbone to model long-range dependencies between regions with similar intensity statistics and produce an initial restored image. In Stage II, a compact residual subtractor processes a 21-channel representation of the degraded input, the Stage-I output, their difference, low- and high-frequency features, and an in-mask prior through a shared convolutional trunk with multiple residual heads, using the direct head to predict a signed residual for final refinement.

\noindent \textbf{Training Details.} The two stages are trained sequentially. Histoformer is initialized from a public checkpoint and fine-tuned on the official UR$^3$ paired data. Its cached outputs are then used to train the Stage-II direct branch for 30K steps on $512\times512$ patches, with a width of 64 and a synthetic-overlay probability of 0.20. The objective combines image reconstruction, residual supervision, multi-band consistency, and clean-region regularization. The pretrained Histoformer checkpoint is counted as external model usage.

\noindent \textbf{Testing Details.} During testing, each degraded image is first processed by the fine-tuned Histoformer. The spatial mask prior is subsequently generated or loaded, and the resulting multi-band representation is fed into the direct residual branch to predict the correction layer in a single forward pass.

\noindent \textbf{Implementation Details.} The Stage-II residual model adopts the selected direct branch with a width of 64. Both stages are integrated into a single deterministic inference pipeline.

\subsection{daniel0902}

This team proposes a lightweight single-image mixture-of-experts network with a shared convolutional backbone, six residual blocks, and 192 base channels. A spatial router activates the top-2 of eight convolutional experts to adaptively handle raindrop- and reflection-degraded regions. The network uses dual reconstruction heads to predict the restored transmission ($T$) and reflection layer ($R$), constrained by the layer-consistency loss ($\lVert (T+R) - I \rVert$). This decomposition enables joint reflection removal and recovery of raindrop-occluded content, while only ($T$) is retained during inference.

\noindent \textbf{Training Details.} The model is trained from scratch on the official training set for 360 epochs using Adam, with $256\times256$ crops, a batch size of 24, and a learning rate of $1\times10^{-4}$. Augmentation includes horizontal flipping and, during continued training, photometric transformations applied with a probability of 0.25, such as exposure and gamma adjustment, color jitter, mild noise, and veiling glare.

\noindent \textbf{Testing Details.} During testing, each image is restored independently using a single checkpoint and full-image inference at \(1080\times720\), without tiling. A two-view flip-based self-ensemble averages the predictions from the original image and its horizontally flipped version.

\noindent \textbf{Implementation Details.} This network contains 10.23M parameters and uses top-2 sparse expert activation. Training is performed on one NVIDIA H100 GPU with mixed precision and takes approximately 9 hours. Inference uses fp16 autocasting, with a reported runtime of approximately 28.5 seconds per image on an RTX 4060 laptop GPU and 1.3 seconds per image on an H100 GPU.

\section*{Teams and Affiliations}
\subsection*{Organizers}

%\noindent\textbf{Title:}
%LoViF 2026 The First Challenge on Unified Removal of Raindrops and Reflections: Methods and Results

\noindent\textbf{Members:}

\noindent
\begin{tabular}{@{}l@{}}
	Zewei He$^{1,2}$ (\texttt{zeweihe@zju.edu.cn}) \\
	Xingyu Liu$^{1,2}$ (\texttt{xingyu\_liu@zju.edu.cn}) \\
	Xi Tong$^{1,2}$ (\texttt{tongxi\_zju@zju.edu.cn}) \\
	Yu Chen$^{1,2}$ (\texttt{yuchen2024@zju.edu.cn}) \\
	Xin Li$^{3}$ (\texttt{xin.li@ustc.edu.cn})
\end{tabular}

\noindent\textbf{Affiliations:}
$^{1}$ Huanjiang Laboratory, Zhuji, China
$^{2}$ Zhejiang University, Hangzhou, China
$^{3}$ University of Science and Technology of China, Hefei, China

\subsection*{Participants}
Please check in our supplementary material.

\section*{Acknowledgment}
This work was supported in part by National Natural Science Foundation of China under Grant No. 52305590, Zhejiang Provincial Natural Science Foundation of China under Grant No. LQ24F010004, and Tianshan Talent Cultivation Plan - Science and Technology Innovation Team Project under Grant No. 2024TSYCTD0011.

\bibliographystyle{splncs04}
\bibliography{main}
\end{document}